\documentclass[10pt,twocolumn,letterpaper]{article}

\usepackage{cvpr}
\usepackage{times}
\usepackage{epsfig}
\usepackage{graphicx}
\usepackage{amsmath}
\usepackage{amssymb}

\usepackage[breaklinks=true,bookmarks=false]{hyperref}

\cvprfinalcopy % *** Uncomment this line for the final submission

\def\cvprPaperID{****} % *** Enter the CVPR Paper ID here

\begin{document}

%%%%%%%%% TITLE
\title{Domain Adaptive Generation of Aircraft on Satellite Imagery\\via Simulated and Unsupervised Learning\thanks{A preliminary version of this paper was presented at \emph{the International Workshop on Machine Learning for Artificial Intelligence Platforms} held in 2017 Asian Conference on Machine Learning (MLAIP@ACML).}}

\author{Junghoon Seo\qquad Seunghyun Jeon\qquad Taegyun Jeon\\
Satrec Initiative\\
Daejeon, South Korea\\
{\tt\small \{sjh, jsh, tgjeon\}@satreci.com}
}

\maketitle
%\thispagestyle{empty}

%%%%%%%%% ABSTRACT
\begin{abstract}
Object detection and classification for aircraft are the most important tasks in the satellite image analysis. The success of modern detection and classification methods has been based on machine learning and deep learning. One of the key requirements for those learning processes is huge data to train. However, there is an insufficient portion of aircraft since the targets are on military action and operation. Considering the characteristics of satellite imagery, this paper attempts to provide a framework of the simulated and unsupervised methodology without any additional supervision or physical assumptions. Finally, the qualitative and quantitative analysis revealed a potential to replenish insufficient data for machine learning platform for satellite image analysis.
\end{abstract}

\section{Introduction}

Satellite image analysis consists of various computer vision and machine learning techniques. In particular, object detection and classification play key roles in the satellite image analysis. For the most part, automatic target recognition has tended to center around the question on machine learning \cite{cheng2016survey}. However, crucial targets are too sparse to be observed because of military operations and actions. For example, the military aircraft as primary targets could be operated rarely in present or be hidden under air-raid shelter.

In order to increase the scarce data onto satellite imagery, numerous studies have attempted to generate synthetic aircraft images. Previous study can be categorized into three approaches: (i) radiometric process \cite{schott1999advanced}, (ii) graphics-based models on optics \cite{ientilucci2003advances}, and (iii) on atmospheric science \cite{han2017efficient}.  Despite the efforts of continuous studies, previous models have limitation to apply for a wide range of situations. Those studies depend on too rigid physical assumption and specific condition, not on real data.

This paper aims to provide an alternative framework for aircraft simulation on satellite imagery. The proposed method would neglect to follow any assumption of physical phenomenon which is hard to be perfectly modeled or appropriately reproduced. %Due to the nature of the satellite imagery, the appearance of the aircraft is rarely deformed. 
During the adversarial training process, the refiner learns how to generate a real-like satellite imagery from synthetic aircraft. %These attempts have not, so far, been noticed, nor have they been studied in detail. 
The remainder of this paper will illustrate the foregoing remarks by considering the simulation model and both qualitative and quantitative results.

\section{Proposed Method}
\begin{figure}[t]
\begin{center}
\includegraphics[width=\linewidth]{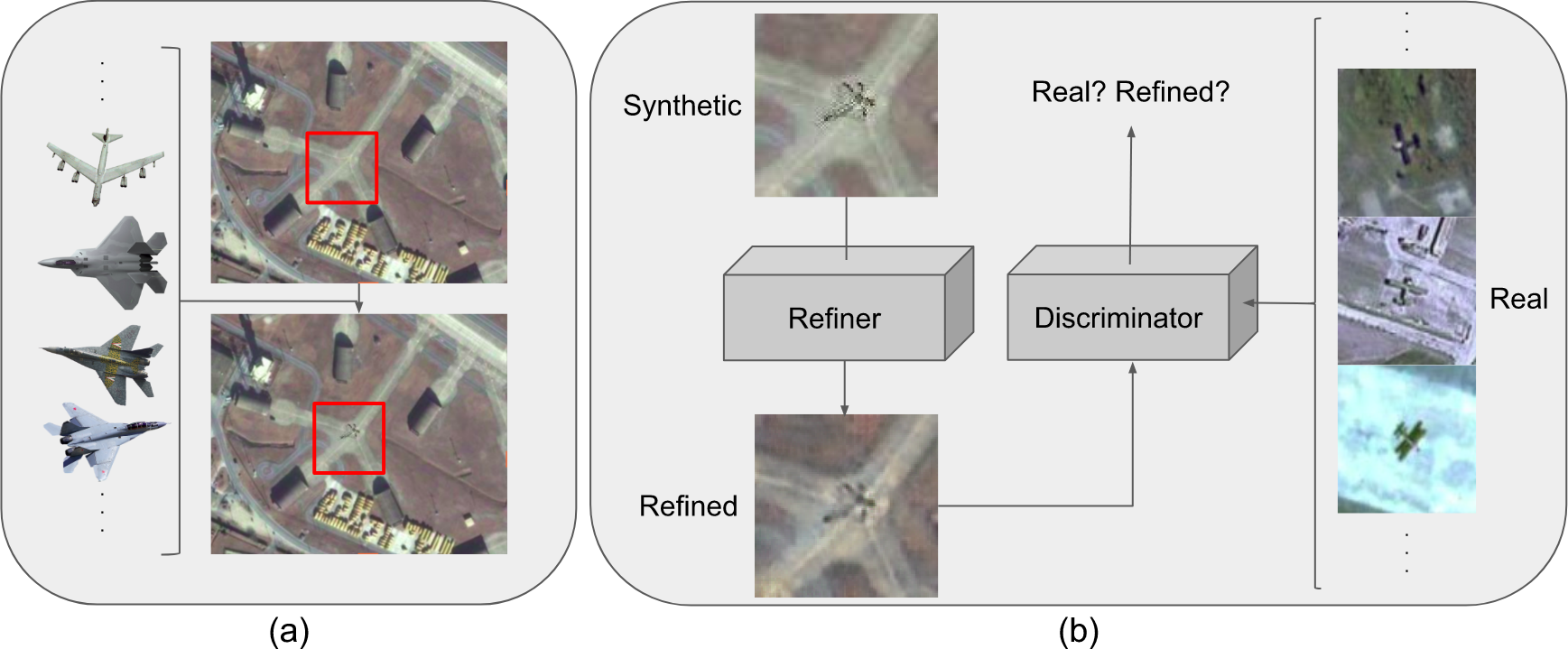}
\end{center}
   \caption{Overview of our proposed method. (a) overlay of aircraft images on satellite imagery. (b) adversarial learning to refine the synthesized image}
\label{pipeline_figure}
\end{figure}

\subsection{Overlay satellite imagery with aircraft images}

Object simulation on the satellite image is treated as a difficult task because of an extremely wide diversity of data rather than object deformation. Moreover, it is usual that quality of images with which the image analysis platform deals is overwhelmingly low. Therefore, we decide not to elaborately concern about a delicate 3D CAD model of aircraft.

This step of our pipeline is shown in Fig. \ref{pipeline_figure} (a). Various near-top-down view images are crawled from web, then crop the background of them. Next, we just overlaid on valid background of a satellite image. 
%These images are much easier to be obtained than CAD models and it is a kind of possible approach because our model does not rely upon any graphical inference. 
These images are much easier to be obtained than CAD models and usage of them makes us free from any bothersome graphical rendering. 
The only augmentation method of aircraft image is a rotation.
% * <taylor.taegyun.jeon@gmail.com> 2017-09-26T00:47:16.314Z:
% 
% > graphical inference
% 이 단어도 의미가 모호합니다. 
% 
% ^.

\subsection{Simulated and Unsupervised Adversarial Learning}

Right after putting aircraft on background, it looks artificial because a visual correlation between object and background is not considered at all. For the harmonious synthesis considering both aircrafts' objectness and background, we adopt a \textit{simGAN} model \cite{shrivastava2017learning} derived from a generative adversarial network \cite{goodfellow2014generative}. Fig. \ref{pipeline_figure} (b) shows that two different neural network models, which are called \textit{refiner} and \textit{discriminator}, are trained in adversarial concept.

Suppose there are two sets of samples $X=\{x_i\}_{i=1}^{n}$ and $Y=\{y_j\}_{j=1}^{m}$ where $x_i\in\mathbb{R}^{\textit{N}}$ and $y_j\in\mathbb{R}^{\textit{N}}$ is sampled from the source domain image and the target domain image, respectively. The goal of the refiner $R_{\theta}:\mathbb{R}^{N}\mapsto\mathbb{R}^{N}$ is to generate the synthetic image $R_{\theta}(x_i)$, which deceives the \textit{discriminator} into classifying it as real image while keeping the pixels as same with $x_i$'s as possible. 
On the other hand, the \textit{discriminator} $D_{\phi}:\mathbb{R}^{N}\mapsto\mathbb{R}$ aims to classify the synthetic image $R_{\theta}(x_i)$ as fake and the real image $y_j$ as real. The overall \textit{refiner} loss $\textit{L}_{R}$ and \textit{discriminator} loss $\textit{L}_{D}$ are defined as follows:
\begin{equation}
{\left \{ \begin{matrix} 
\quad\!\!\textit{L}_{R} = -$$\sum_{i} \log(1-D_{\phi}(R_{\theta}(x_i))) + \lambda\left \| R_{\theta}(x_i) - x_i \right \|_{1} \\
         \textit{L}_{D} = -$$\sum_{i} \log(D_{\phi}(R_{\theta}(x_i))) -$$\sum_{j} \log(1-D_{\phi}(y_j))
\end{matrix} \right.}
\end{equation}
where $\lambda\in\mathbb{R}$ is a hyper-parameter of the weights for the identity mapping. By using gradient descent method in training step, $\theta$ and $\phi$ are updated alternately to minimize $\textit{L}_{R}$ and $\textit{L}_{D}$, respectively. Finally, a set of the refined images $\hat{X}=\{R_{\theta}(x_i)\}_{i=1}^{n}$ could be generated from the \textit{refiner}, which is real-like but similar to the original $X$. In our task, $X$ is a set of the synthetic images fake aircraft are overlaid on, and $Y$ is a set of the real images authentic aircraft appear on.
% * <taylor.taegyun.jeon@gmail.com> 2017-09-26T01:56:09.199Z:
% 
% > In our task, $X$ is a set of the synthetic images object image overlays on and $Y$ is a set of the real images aerial object appears on.
% 문장 의미 확인
% 
% ^.

\section{Experiments and Discussions}
\begin{figure}[t]
\begin{center}
\includegraphics[width=\linewidth]{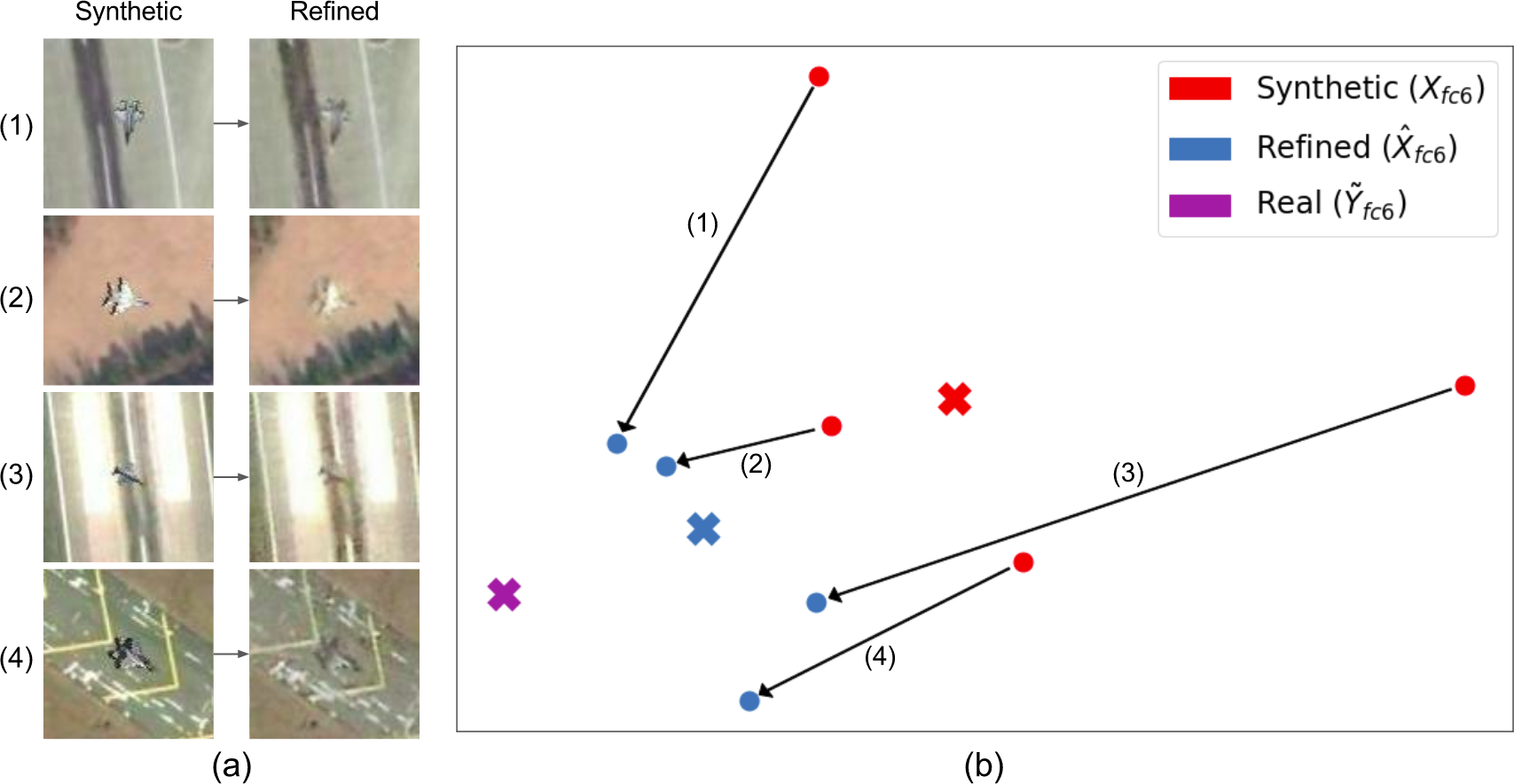}
\end{center}
   \caption{Results from our method and their visualization on t-SNE. (a) Four synthetic samples before and after refinement. (b) Visualization on t-SNE. 'Ex' mark ($\times$) refers mean of embedded manifold of each set (($X_{fc6}$, $\hat{X}_{fc6}$, $\tilde{Y}_{fc6}$)). 'Circle' ($\bullet$) and 'arrow' ($\rightarrow$) marks signify the example images of (a) and matchings between the synthetic and refined pair, respectively.}
\label{qual_figure}
\end{figure}

\subsection{Experiment Details}
We collected RGB satellite imagery using \textit{Google Earth Pro} 7.1 \cite{gorelick2017google}. The dataset includes 8,604 real satellite image patches which include at least one combat aircraft, and 2,917 fake image patches which include overlaid near-top-down aircraft image. All aircraft models of the overlaid images are totally different from those of the collected images. %The size of patch is 96$\times$96. The resolution is about 1.4 meter per pixel.

We substitute our neural network architecture with a one-way version of \cite{Zhu2017cycle}, which is said that its architecture is appropriate to solve style transfer task. $\lambda$ is set as $40$. On training step, batch size is one and training is over after 180k step. In evaluation step, $\tilde{Y}=\{\tilde{y_k}\}_{k=1}^{2917}$ is sampled independently and identically from $Y$ without replacement. Setting of the other hyperparameters is the almost same as those in \cite{shrivastava2017learning}.

\subsection{Qualitative Evaluation}

Refinement results are shown in Fig. \ref{qual_figure} (a). After the refinement, the aircraft look much more natural than those in the original synthetic images. To examine the visual results analytically, we apply t-SNE \cite{maaten2008tnse}. We select pre-trained VGG19 \cite{simonyan2015very} which is already proven to be useful in neural style transfer task \cite{gatys2016image}. \textit{fc6} layer feature vectors $X_{fc6}$, $\hat{X}_{fc6}$, and $\tilde{Y}_{fc6}$ are extracted from $X$, $\hat{X}$, and $\tilde{Y}$, respectively.

Fig. \ref{qual_figure} (b) shows a visualization of the result from t-SNE. The average of $\hat{X}_{fc6}$ is closer to $\tilde{Y}_{fc6}$ than to $X_{fc6} $ and it makes roughly a conjecture that the domain difference between the refined and the real is smaller than that between the synthetic and the real. Additionally, it appears consistent that each example is shown as a point also moves closer to the distribution of the real image after the refinement.

\subsection{Quantitative Evaluation}
Maximum mean discrepancy (MMD) \cite{gretton2012kernel} is one of test statistics for measurement of the difference between two distributions. A mixture of 16 Gaussian radial basis functions (RBF) kernels, where sigma varies from $10^{-6}$ to $10^6$, is considered to be associated continuous kernel. To avoid excessive ${O}(n^2\log n)$ or ${O}(n^3)$ time complexity, a linear time unbiased estimate of MMD is used \cite{gretton2012kernel}.

\begin{table*}[tbp]
\centering
\caption{Comparisons of maximum mean discrepancy between each image pair}
\label{mmd_table}
\begin{tabular}{c|c|c}
\hline
(i) $MMD(X_{fc6}, \hat{X}_{fc6})$ & (ii) $MMD(X_{fc6}, \tilde{Y}_{fc6})$ & (iii) $ MMD(\hat{X}_{fc6}, \tilde{Y}_{fc6})$ \\ \hline
0.3329                        & 0.3423                          & \textbf{0.2300 }                               \\ \hline
\end{tabular}
\end{table*}

We report three MMD values among $X_{fc6}$, $\hat{X}_{fc6}$, $\tilde{Y}_{fc6}$ in Table \ref{mmd_table}. It is worthy to note that (iii) is the smallest. It makes us infer that domain difference between $\tilde{X}$ and $Y$ is even lower than that between the other pairs, i.e. $X$ and $\tilde{X}$, or $X$ and $Y$. The result shows that the proposed method effectively reduces the gap between synthetic image and real image in terms of quantitative analysis.

% We report MMD value among $X_{fc6}$, $\hat{X}_{fc6}$, $\tilde{Y}_{fc6}$ in Table \ref{mmd_table}. Compared to $MMD(X_{fc6}$, $\tilde{Y}_{fc6})$, $MMD(\hat{X}_{fc6}$, $\tilde{Y}_{fc6})$ is quiet small. It makes us infer that domain difference between $\tilde{X}$ and $Y$ is a lot smaller than that between {X} and {Y}. Even comparing $MMD(\hat{X}_{fc6}, \tilde{Y}_{fc6})$ with $MMD(X_{fc6}, \hat{X}_{fc6})$, it implies the same argument. $MMD(\hat{X}_{fc6}, \tilde{Y}_{fc6})$ value itself is enough low to be regarded impressive, considering that the airplane models in $\hat{X}$ are completely different from the models in $\tilde{Y}$.

\section{Conclusions}

In this paper, we introduced our method to build up the graphics-free simulation for aircraft in satellite imagery. Our approach is based on data-dependent simulated and unsupervised learning method so it could be freely adaptable to any condition of the similar tasks. The experiment shows meaningful qualitative and quantitative performance. In future work, we will focus on improving the performance of the classification and detection method of our satellite image analysis platform using refined image data. We expect that our method will contribute to the field of remote sensing, especially in data generation for automatic target recognition.
{\small
\bibliographystyle{ieee}
\bibliography{egbib}
}

\end{document}